\documentclass[runningheads]{llncs}
\usepackage{graphicx}

\usepackage{tikz}
\usepackage{comment}
\usepackage{amsmath,amssymb} 
\usepackage{color}

\usepackage[accsupp]{axessibility}  

\usepackage[width=122mm,left=12mm,paperwidth=146mm,height=193mm,top=12mm,paperheight=217mm]{geometry}

\usepackage{booktabs}
\usepackage{enumitem}
\usepackage{epsfig}
\usepackage[pagebackref,breaklinks,colorlinks]{hyperref}

\usepackage{adjustbox}
\usepackage{xspace}
\makeatletter
\DeclareRobustCommand\onedot{\futurelet\@let@token\@onedot}
\def\@onedot{\ifx\@let@token.\else.\null\fi\xspace}

\def\aka{\emph{a.k.a.}}
\def\eg{\emph{e.g}\onedot} 
\def\ie{\emph{i.e}\onedot} 
 
\def\etc{\emph{etc}\onedot} 
\def\wrt{w.r.t\onedot}

\makeatother

\begin{document}
\pagestyle{headings}
\mainmatter
\def\ECCVSubNumber{5396}  

\title{Motion Transformer for Unsupervised Image Animation} 


\titlerunning{Motion Transformer for Unsupervised Image Animation}
%

\author{Jiale Tao\inst{1}$^{*}$\and
Biao Wang\inst{2}\and
Tiezheng Ge\inst{2}\and
Yuning Jiang\inst{2}\and
Wen Li\inst{1}$^{\dagger}$\and\\
Lixin Duan\inst{1}}
\authorrunning{Tao et al.}
%
\institute{School of Computer Science and Engineering \& Shenzhen Institute for Advanced Study, University of Electronic Science and Technology of China\\
\email{\{jialetao.std, liwenbnu, lxduan\}@gmail.com}\\
Alibaba Group\\
\email{\{eric.wb, tiezheng.gtz, mengzhu.jyn\}@alibaba-inc.com}}

\maketitle

\renewcommand{\thefootnote}{\fnsymbol{footnote}}
\footnotetext{$*$ Work done during an intership at Alibaba Group, code is available at https://github.com/JialeTao/MoTrans \\$\dagger$ Corresponding author}

\begin{abstract}
Image animation aims to animate a source image by using motion learned from a driving video. Current state-of-the-art methods typically use convolutional neural networks (CNNs) to predict motion information, such as motion keypoints and corresponding local transformations. However, these CNN based methods do not explicitly model the interactions between motions; as a result, the important underlying motion relationship may be neglected, which can potentially lead to noticeable artifacts being produced in the generated animation video. To this end, we propose a new method, the motion transformer, which is the first attempt to build a motion estimator based on a vision transformer. More specifically, we introduce two types of tokens in our proposed method: i) image tokens formed from patch features and corresponding position encoding; and ii) motion tokens encoded with motion information. Both types of tokens are sent into vision transformers to promote underlying interactions between them through multi-head self attention blocks. By adopting this process, the motion information can be better learned to boost the model performance. The final embedded motion tokens are then used to predict the corresponding motion keypoints and local transformations. Extensive experiments on benchmark datasets show that our proposed method achieves promising results to the state-of-the-art baselines. Our source code will be public available.

\end{abstract}

\section{Introduction}

Image animation (also known as motion transfer) is a technique that aims to animate a source image based on the motion information extracted from a given driving video, such that the generated video can mimic the motion in the driving video while simultaneously retaining the appearance of the target object in the source image.
This approach enables people to quickly create innovative content without the need to start from scratch, which can save large amounts time.
Motion transfer has gained significant attention from the computer vision community in recent years~\cite{liu2019liquid,chan2019everybody,geng20193d,nirkin2019fsgan,siarohin2019animating,siarohin2020first,PCAMotion}, owing to its wide range of practical applications across entertainment and education such as virtual try-on~\cite{bhatnagar2019multi,chopra2021zflow}, video conferencing~\cite{wang2021one}, e-commerce advertising~\cite{xu2021move}, and so on.

Existing image animation works can be roughly divided into two categories, namely supervised methods and unsupervised methods. In more detail, supervised methods typically focus on the animation of a specific object type (\eg, human body, human face, \etc) and utilize a third-party model to extract structural representations, which might take the forms of 2D keypoints~\cite{ma2017pose,ma2018disentangled}, 3D meshes~\cite{zhao2018learning}, 3D optical flow~\cite{liu2019liquid} and so on. This type of method has advantages in modeling accurate object structures, but is limited by the object-specific approach to image animation. On the other hand, unsupervised methods~\cite{siarohin2019animating,siarohin2020first,PCAMotion} aim to avoid the requirement for object-specific predefined structure representations. These approaches usually learn intermediate motion representations (\eg, keypoints and affine matrices) between two images by warping one image to reconstruct another. Currently proposed unsupervised methods generally comprise of two modules: a motion estimator and an image generator. In these methods, the image generators tend to be quite similar,  while the motion estimators are always the research focus and are proven to be quite crucial for animation performance. For example, Siarohin et al.~\cite{siarohin2019animating} utilize an unsupervised keypoint detector to estimate sparse motions. These authors later boost the performance of their method by adding a head in order to better predict the affine matrix~\cite{siarohin2020first}. Moreover, the method proposed in~\cite{PCAMotion} further improves the motion learning process, by entangling a keypoint and the corresponding affine matrix into a single heat-map estimation.

In order to animate arbitrary objects, we follow the unsupervised setting and focus primarily on motion estimation in this work. It is worth noting that all CNN-based methods discussed above fail to consider the interactions between motions, which may prevent these methods from learning robust motion estimators. We believe the robustness of motion estimators can be boosted by the global information of motions. Accordingly, in this work, we make the first attempt to model the global motion information by employing vision transformers in unsupervised image animation. More specifically, we explicitly model the motion (\ie, the keypoint and corresponding affine matrix) as a query token in the transformer, which we refer to as motion tokens and treat as learnable parameters. We further introduce image tokens, which are obtained by projecting the flattened image patch features to the same dimension as the motion tokens. These motion tokens, conditioned on image tokens, are then decoded to final keypoints and affine matrices through several transformer layers. Intuitively, the motion transformer compensates for the lack of prior structural representations by naturally introducing global motion information to assist with part motion learning; this procedure is efficiently implemented through the self-attention mechanism. We can summarize the advantages of the motion transformer in two aspects with reference to different objects: i) For objects with relatively non-rigid motions (such as human body), it learns the set of local motions in a more stable fashion; ii) for objects with relatively rigid motions (such as faces), it exhibits a strong ability to learn global motion patterns simultaneously for all motion tokens.

We conduct extensive experiments on four benchmark datasets, which contain various kinds of objects such as talking heads, human bodies, animals, etc. The superior performance of our proposed method relative to existing baselines clearly demonstrates that global motion information can help to improve the robustness of motion estimators, as well as showing the success of our proposed motion transformer in capturing the global motion information.

\section{Related Work}

\noindent{\bf Image animation:} Supervised methods~\cite{ma2017pose,ma2018disentangled,geng20193d,nirkin2019fsgan,ren2020human,zakharov2019few,zhu2019progressive,li2019dense,chan2019everybody,ren2021flow,zhang2021pise,siarohin2018deformable,neverova2018dense} focus on the animation of a specific object type. Among these, the human body \cite{lorenz2019unsupervised,ren2020human,huang2021few,dwnet2019,ren2020deep,chen2019unpaired,balakrishnan2018synthesizing,Sarkar2020,Yoon_2021_CVPR,jiang2022text2human} and human face~\cite{geng20193d,ha2020marionette,pumarola2018ganimation,Yao_2021,Wei_2021,wiles2018x2face,burkov2020neural,tripathy2021facegan,wei2020learning,kim2018deep} are the most popular animation objects. Methods of this kind rely on object-specific landmark detectors, 3D models or other forms of supervision, which are usually pre-trained on a large amount of labeled data. On one hand, the advantage of these methods is that based on the pre-obtained structure representations, it is easier to further learn the warping flow between two images. On the other hand, these methods are also hampered by an obvious limitation, as they are only suitable for a specific object type. 

Unsupervised methods~\cite{siarohin2019animating,siarohin2020first,PCAMotion,tao2022structure} have been recently proposed to address the above limitation. These approaches typically leverage a large amount of easy-to-obtain unlabeled web videos and design image reconstruction losses to learn intermediate motion representations (\eg, keypoints and affine matrices). Benefiting from the unsupervised scenario, methods of this kind can be applied to animate a wide range of objects, including human bodies, human faces, animals, etc. It is worth noting that no predefined structural representations of objects are available for training in those methods. 
Specifically, Monkey-Net~\cite{siarohin2019animating} proposes to learn intermediate part keypoints as sparse motions by means of the downstream image reconstruction task. Subsequently, FOMM~\cite{siarohin2020first} improves on this approach by simultaneously regressing the local affine matrices along with the keypoints of object parts. Moreover, MRAA~\cite{PCAMotion} further improves FOMM by combining the learning of part keypoints and local affine matrices into a single heat-map estimation process. 
Among these approaches, however, Monkey-Net is limited by the coarsely defined motion model, FOMM suffers from regressing stable affine matrices, while MRAA struggles in modeling relatively rigid motions (\eg, human face) and fails to consider cooperative part motions. 

It is worth noting that all above unsupervised methods focus on the motion estimation process in image animation. Our method also lies in this research scope, with a newly proposed motion transformer as the motion estimator. Similar to FOMM, our method also regresses the affine matrices, while it avoids the instability problem by adopting a global-assisted approach; moreover, benefiting from this approach, our method is also better able to handle rigid motions and cooperative local motions when compared with MRAA.

\noindent{\bf Vision transformer:} Transformers~\cite{vaswani2017attention} have achieved great success in natural language processing (NLP) community. Recently, they have also achieved promising results in computer vision tasks. Among them, DETR~\cite{carion2020end} and ViT~\cite{dosovitskiy2020image} are the pioneering methods, and have been followed by a series of other vision transformer methods~\cite{zhu2020deformable,zheng2021rethinking,li2021diverse,li2021tokenpose,yu2021pointr,yang2021transpose,wan2021high}. DETR~\cite{carion2020end} was the first to introduce transformers to the object detection. 
It follows the encoder-decoder architecture of traditional transformers in NLP, where object queries are introduced as learnable parameters. ViT~\cite{dosovitskiy2020image} proposes a transformer encoder architecture for image classification, which directly splits the image into patches and introduces a learnable classification token to aid in performing the task. Recently, several methods have proposed variant forms of transformers for landmark detection~\cite{watchareeruetai2021lotr,li2021tokenpose,yang2021transpose}. Our method is motivated by these recent works, in that we regress the object keypoints as well as affine matrices for image animation. 

\section{Methodology}

In the context of unsupervised image animation, we are given a source image $S$ and a driving video $D=\{Z_i\}$, where $Z_i$ is the $i$-th video frame. Unless otherwise noted, in the remainder of this work we will use $Z$ to represent a video frame for the sake of simplicity. 

\subsection{The General Framework for Image Animation }\label{3.1}
Existing unsupervised image animation methods~\cite{siarohin2019animating,siarohin2020first,PCAMotion} generally perform image animation in a frame-by-frame manner. Given the source image and each frame of the driving video, the image animation model outputs a synthesized image that mimics the pose of the object in the driving frame while also preserving the appearance of the object in the source image. 

Unsupervised models typically comprise two stages: motion estimation and image generation. The motion estimation stage produces the relative motion (often in the form of optical flow) between each driving video frame and the source image, while the image generation stage warps the source image based on the relative motion in order to generate the synthesized image. 

To obtain the relative motion, the motion information of the source image or a driving video frame is first separately predicted and then ensembled to calculate the dense motion flow. More specifically, the motion information of a single image is disentangled as a set of transformations of object parts, each of which is represented by a keypoint and its affine transformation from an latent reference image. A motion estimator is designed to predict the keypoints and the corresponding affine transformations for the input image.

\begin{figure*}[ht]
\centering
  \includegraphics[width=1.0\linewidth]{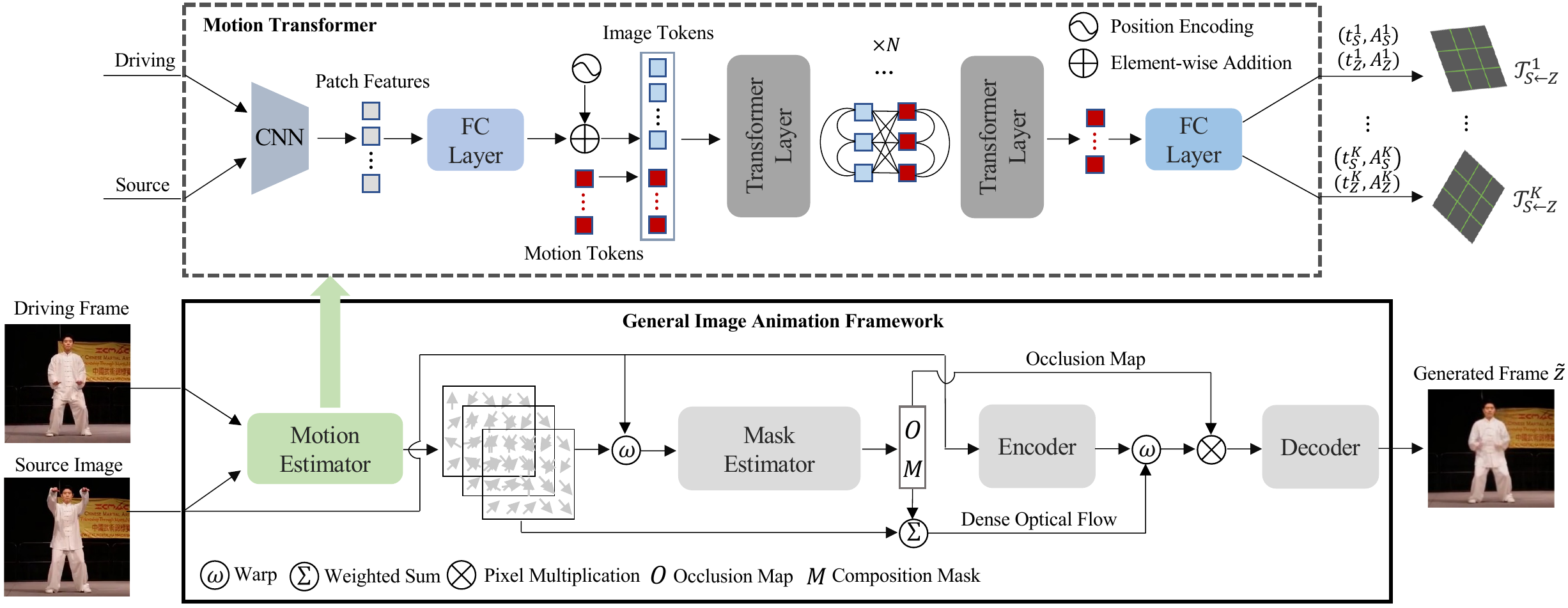}
  \caption{Overview of the general image animation framework and our proposed motion transformer. Unlike the existing CNN based works~\cite{siarohin2020first,PCAMotion}, our motion transformer introduces image tokens and motion tokens, which encode visual and motion information respectively. And those tokens are further sent into multiple transformer layers to mine the underlying interactions between them, the self attention and cross attention are denoted by the straight and curved lines. By using a linear head, the output motion tokens are finally regressed to keypoints and their corresponding affine matrices.}
  \label{fig:pipeline}
\end{figure*}

\noindent\textbf{Motion estimation:}
The first stage of the general image animation framework involves estimating relative motions between the source image and driving frame. This stage plays a critical role in the process, as the estimation accuracy largely determines the overall quality of the generated video. Existing works~\cite{siarohin2019animating,siarohin2020first,PCAMotion} generally follow CNN-style models, where the transformation of each object part is derived by learning a mapping from the learned feature maps. We contend that current CNN-based methods may not adequately capture the global motion information, as they do not consider interactions between part motions.

Assume there are $K$ parts in an object for either the source image or each driving frame. In the motion estimation stage, the goal is to learn a motion transformation $(t^k, A^k)$ for the $k$-th part, where $t^k\in \mathbb{R}^{2\times 1}$ denotes a keypoint (\ie, the centroid of the transformation), $A^k\in \mathbb{R}^{2\times 2}$ represents the corresponding affine transformation matrix, and $k=1,...,K$.

Moreover, since the affine transformation $A^k$ should be applied only to a certain neighboring area (also known as a mask) of the object part rather than the entire image, we constrain the effect of $A^k$ by further learning the corresponding mask $M^k$. In the literature~\cite{siarohin2020first,PCAMotion}, the mask estimator is usually designed as a CNN-based encoder-decoder architecture. Specifically, it takes the warped source image as input and generates the masks $\{M^k|_{k=1}^K\}$'s of $K$ object parts. Furthermore, an additional occlusion map can also be learned to guide the image generator in inpainting the occluded regions. We refer the readers to~\cite{siarohin2020first,PCAMotion} for further details.

\noindent\textbf{Motion representation:} After learning $(t^{k}_S, A^{k}_S)$ and $(t^{k}_Z, A^{k}_Z)$, we can obtain the following motion flow from the driving frame $Z$ to the source image $S$ for the $k$-th object part based on the first-order motion model~\cite{siarohin2020first,PCAMotion}, as follows:
\begin{equation}
\mathcal{T}^k_{S\leftarrow Z}(c) = t^k_S + A^k_S(A^k_Z)^{-1}(c-t^k_Z), \label{FOMM_local}
\end{equation}
where $c$ denotes any image coordinate in the driving frame. 

We next obtain the dense motion flow $\mathcal{T}_{S\leftarrow Z}(c)$ by combining $\mathcal{T}^k_{S\leftarrow Z}(c)$ with masks $M^k(c)$ as linear weights:
\begin{equation}
\mathcal{T}_{S\leftarrow Z} (c) = \sum_{k=1}^{K}M^k(c)\cdot\mathcal{T}^k_{S\leftarrow Z}(c), \label{FOMM}
\end{equation}
where $M^k(c)$ is the mask of the $k$-th object part at the coordinate $c$ and $\sum_{k=1}^{K}M^k(c)=1$. Moreover, the dense motion flow $\mathcal{T}_{S\leftarrow Z}(c)$ represents that the pixel value at coordinate $c$ of the generated image $\tilde{Z}$ is warped and obtained based on the pixel value at coordinate $\mathcal{T}_{S\leftarrow Z}(c)$ of the source image.

\noindent\textbf{Image generation:} Given the source image $S$ and the dense motion flow $\mathcal{T}_{S\leftarrow Z}(c)$, a Unet-based encoder-decoder generator is introduced to generate the synthesized image $\tilde{Z}$. Specifically, the source image $S$ is passed through the encoder to obtain feature maps, after which it is then warped according to the dense motion flow $\mathcal{T}_{S\leftarrow Z}(c)$. Finally, the decoder learns the synthesized image $\tilde{Z}$ based on the warped feature map.

\subsection{Motion Transformer}\label{3.2}
Since existing CNN-based methods do not explicitly model the interactions between motions, the underlying motion relationship is not fully exploited and cannot be properly captured. We argue that this underlying relationship is critical to the process and helps reduce artifacts in generated animation videos. For instance, when people smile, the movement of their mouths and eyes occurs simultaneously, meaning that they are highly correlated. Considering this limitation of existing CNN-based models, we aim to seek out a better way of modeling the motion interactions.

To address the above issue, we propose to take advantage of the recently proposed vision transformer. We accordingly name our method the {\it motion transformer}. In a vision transformer layer, raw data are processed to form tokens, which act as the layer input. The underlying relationship among those tokens can be effectively mined through the attention mechanism.
As a result of adopting this approach, meaningful embeddings can be learned for those tokens. 
Our motion transformer employs multiple vision transformer layers.

In our proposed motion transformer for image animation, we explicitly model the motions of object parts as input query tokens ({\it motion tokens}) to the transformer. We further obtain {\it image tokens} by projecting image patch features through a fully connected layer and subsequently embedding them with position encoding. By feeding those two types of tokens together into the transformer, the motion tokens are able to utilize the global context information of the entire image through attention with image tokens, which aids in better capturing the interaction between object part motions. Moreover, a linear head is designed in the last transformer layer to directly regress the keypoints and affine matrices of the motions. The entire process is illustrated in Fig~\ref{fig:pipeline}. 

\noindent\textbf{Tokens:} Two types of tokens are introduced in our motion transformer. We first introduce a set of motion tokens, inspired by the recent vision transformers~\cite{carion2020end}. Each motion token is expected to encode the motion information of an object part (\ie, a keypoint and its affine transformation). These motion tokens are considered as learnable embeddings in our method; we denote them as $\{P^k_0|^K_{k=1}\}$, where $P^k_0\in \mathbb{R}^d$ represents the $k$-th object part and $d$ is the embedding dimension. 

The second type of tokens is the image token. Rather than directly using raw images, we first extract low-level image features by utilizing a CNN model. Subsequently we flatten the patch image features, with each patch projected to dimension $d$. To maintain the position information of the patch image features, we add the projected features by the absolute position encoding. We refer to the result features as image tokens, denoted as $I^n_0\in \mathbb{R}^d, n=1,...,N$.


\noindent\textbf{Multiple vision transformers:}
When an object moves, different object parts are not completely independent. Rather, they often correlate with each other, which, however, was not discussed in existing motion transfer methods~\cite{siarohin2020first,PCAMotion}. To model the relation among tokens, we utilize the natural advantages of vision transformer for building attention. In particular, a motion token is updated via all motion tokens and image tokens, and correspondingly we build two types of attention for the motion tokens. i) self attention for mining the underlying relationship between motion tokens; ii) cross attention for decoding motion tokens to the final keypoints and affine matrices. 

Formally, let us denote by $P^i_{l-1}$ a motion token that to be input to the $l$-th transformer layer, in which  $P^i_{l-1}$ is linearly projected to the query, key and value features $Q_{P^i_{l-1}},K_{P^i_{l-1}},V_{P^i_{l-1}}$; and similarly for image tokens we have $Q_{I^i_{l-1}},K_{I^i_{l-1}},V_{I^i_{l-1}}$. For ease of illustration, we temporally drop the subscript and define the multi-head self attention ($\operatorname{MSA}$) as follow:
\begin{equation}
    head^j=\operatorname{softmax}\left (\frac{QW^j_{Q}\left (KW^j_K\right)^T}{\sqrt{d}}\right)VW^j_V,
\end{equation}
\begin{equation}
   \operatorname{MSA}\left (Q,K,V\right)=[head^{1},...,head^{h}]W_O,
\end{equation}
$W_O,W^j_Q,W^j_K,W^j_V$ are learnable parameters, and $h$ represents the total number of heads in each transformer layer. In practice, $Q$ is the query from a token, while $K,V$ are from another token. With this definition, the motion token is updated by self attention and cross attention as follows:
\begin{equation}
P^i_l=\sum_j \operatorname{MSA}(Q_{P^i_{l-1}},K_{P^j_{l-1}},V_{P^j_{l-1}})+\sum_j \operatorname{MSA}(Q_{P^i_{l-1}},K_{I^j_{l-1}},V_{I^j_{l-1}}),
\label{self_cross_att}
\end{equation}
\begin{equation}
    P^i_l =\operatorname{FFN}\left (\operatorname{LN}\left (P^i_l\right)+P^i_l\right),
\end{equation}
where $\operatorname{FFN}$ and $\operatorname{LN}$ denote the feed forward network and layer normalization respectively. On one hand, the left term of Eqn.~(\ref{self_cross_att}) (\ie, the self attention) indicates that each motion token tends to query all other motion tokens, in this way the underlying relationship between motion tokens could be effectively captured; on the other hand, the motion token can be gradually embedded with motion information through querying the image tokens, as formulated in the right term of Eqn.~(\ref{self_cross_att}) (\ie, the cross attention). 


The two types of attention process above enable the efficient interactions between tokens. In implementation, different from recent works~\cite{carion2020end,li2021diverse} in which the two types of attention process are separately conducted, we found it more efficient to unify the self attention and cross attention in a single transformer architecture. To do this, we directly concatenate the image tokens and motion tokens as the initial input tokens $F_0=[P^1_0;...;P^K_0;I^1_0;...;I^N_0]\in \mathbb{R}^{(N+K)\times d}$, and use the single self attention process to update the concatenated tokens:
\begin{equation}
F^i_{l}=\sum_j \operatorname{MSA}(Q_{F^i_{l-1}},K_{F^j_{l-1}},V_{F^j_{l-1}}),
\end{equation}
\begin{equation}
    F^i_{l}=\operatorname{FFN}\left (\operatorname{LN}\left ( F^i_{l}\right)+F^i_{l}\right).
\end{equation}

Note that in this procedure, image tokens are also updated with attention to all tokens, while to our observation, this dose not affect the performance, and the unified self attention is more efficient in end-to-end training.

At the final transformer layer, we take out the motion tokens $P_L^k$ from the output token feature $F_L$, and employ a linear head to directly regress the affine matrix and translation vector. Let $W_h\in \mathbb{R}^{d\times 6}$ denote the parameters of the linear head; the decoded part affine matrix and translation vector of each motion token are then computed as $[A^k,t^k]=P_L^kW_h$.

\subsection{Training} \label{3.3}
We consider the following losses to formulate the objective function of our method. The whole training process is conducted in an end-to-end fashion.

\noindent\textbf{Perceptual loss:} Following FOMM and MRAA, we adopt the multi-resolution perceptual loss~\cite{johnson2016perceptual} defined with a pre-trained VGG-19~\cite{simonyan2014very} network. Given the driving frame $Z$ with resolution index $i$, the generated image $\tilde{Z}$, and the feature extractor $\phi$ with layer index $l$, the perceptual loss can be written as follows:
\begin{equation}
    \mathcal{L}_{per}=\sum_i\sum_{l}\left\|\phi_{l}\left(Z_i\right)-\phi_{l}(\tilde{Z_i})\right\|_1.
\end{equation}

\noindent\textbf{Equivariance loss:} Following FOMM and MRAA, the equivariance loss is adopted here. Given a random geometric transformation $\mathbf{T}$ and a driving image $Z$, this loss can be written as follows:
\begin{equation}
    \mathcal{L}_{equi}=\sum_{k}\left\|\mathbf{T} (t_{Z}^k)-t_{\mathbf{T}(Z)}^k\right\|_1.
    \label{equi loss}
\end{equation}

\noindent\textbf{Background losses:} To handle those situations in which the background is not static, we follow the MRAA in utilizing a background predictor network to predict the background motion flow. To facilitate the separation of the background and foreground motion learning, inspired by~\cite{Gao2021ucos}, we adopt the following losses:
\begin{equation}
    \mathcal{L}_{mask}=\left\|M^{0}-1\right\|_{1}+\sum_{k \neq 0}\left\|M^{k}-0\right\|_{1},
    \end{equation}
and
\begin{equation}
    \mathcal{L}_{con}=\sum_{k \neq 0} \sum_{c} M^{k} (c) \cdot\left (c-u^{k}\right)^{2} / \operatorname{sum}\left(M^{k}\right),
\end{equation}
where $u^k=\sum_{c} M^{k} (c)\cdot c / \operatorname{sum}\left(M^{k}\right)$. Intuitively, the mask loss $\mathcal{L}_{mask}$ is used to constrain the portion of foreground and background area in an image, and the foreground motion mask is considered to be more focused under the constraint of the concentration loss $\mathcal{L}_{con}$.

\noindent\textbf{Overall loss:} We combine all of the above losses to formulate the overall object function of our proposed vision transformer, as follows:
\begin{equation}
    \mathcal{L} = \mathcal{L}_{per} + \mathcal{L}_{equi} + \lambda(\mathcal{L}_{mask}+\mathcal{L}_{con}) .
    \label{total loss}
\end{equation}
It should be noted here we adopt only the above background losses $\mathcal{L}_{mask}$ and $\mathcal{L}_{con}$ for the TaichiHD dataset with $\lambda=0.1$ to handle the dynamic background change in TaichiHD videos in the experiments. As videos from other datasets typically have a static background, we omit $\mathcal{L}_{mask}$ and $\mathcal{L}_{con}$ from the overall loss for those datasets.

\section{Experiments}


\noindent\textbf{Datasets:} The following benchmark datasets are used in our experiments:
\begin{itemize}[itemsep=0pt]
    \item{VoxCeleb~\cite{nagrani2017VoxCeleb}:  A talking head dataset consisting of 20047 videos. 
    All videos are cropped and resized to $256\times 256$.} 
    \item{TaiChiHD~\cite{siarohin2020first}: This dataset contains 3120 videos.
    All videos are cropped and resized to $256\times 256$.}
    \item{TED-talks~\cite{PCAMotion}: This is a talking show dataset containing 1255 videos. 
    All videos are cropped and resized to $384\times 384$.}
    \item{MGIF~\cite{siarohin2019animating}: This dataset contains 1000 cartoon animal videos, 
    all of which are resized to $256\times 256$, following~\cite{siarohin2020first}.} 
\end{itemize}

\noindent\textbf{Evaluation metrics:} We follow \cite{siarohin2020first,PCAMotion} in evaluating the video reconstruction quality, where videos are reconstructed with appearance representations by using their first frame and motion representations \wrt all frames. Four commonly used evaluation metrics are listed below.
\begin{itemize}[itemsep=0pt]
    \item{L1 distance: The average L1 distance between the generated and ground-truth video frames.}
    \item{Average keypoint distance (AKD): The average distance of detected keypoints between the generated and ground-truth video frames. This metric is designed for evaluating the pose quality of generated videos.}
    \item{Missing keypoint rate (MKR): The percentage of keypoints that are not detected in the generated video frames but do exist in the ground-truth. 
    }
    \item{Average Euclidean distance (AED): The average Euclidean distance between generated and ground-truth video frames, as in the feature space. This metric evaluates the identity information of the generated video frames.}
\end{itemize}

\noindent\textbf{Implementation details:} For image generation, we adopt Unet~\cite{ronneberger2015u} to construct the mask predictor and generative encoder-decoder. Skip connections are added in the encoder-decoder architecture similar to~\cite{siarohin2019animating,PCAMotion}. For motion estimation, we adopt the first three stages of the HRNet-W32 encoder~\cite{sun2019deep} pretrained on ImageNet~\cite{deng2009imagenet} as the CNN backbone in our model. After the CNN encoder has been applied, image features are down-sampled by a scale factor of 4. We utilize a 12-layer standard transformer encoder architecture. Moreover, the sine function~\cite{vaswani2017attention} is used for position encoding. 
The image patch size is set to $4\times 4$ in all experiments, with 256 image tokens used for the input resolution of $256\times 256$ and 576 for $384\times 384$. The token dimension $d$ is set to $192$. The number of motion tokens is set to 10, as in~\cite{siarohin2020first,PCAMotion}. The Adam optimizer~\cite{kingma2014adam} is adopted, where the initial learning rate is set as $2\times10^{-4}$ and dropped by a factor of 10 at the end of $60th$ and $90th$ epoch. 
We train the entire networks on eight NVIDIA V100 GPU cards for 100 epochs. 

\subsection{Comparison with State-of-the-Art}

\noindent\textbf{Model capacity:} Under the general image animation framework, we analyze the difference between motion estimators from the model capacity view. As listed in Table~\ref{params_gflops}, the proposed motion estimator has slightly less parameters than FOMM and MRAA, while the FLOPs of it are much heavier, which is caused by the high-resolution (4 times lower than the input image resolution) computation in the CNN encoder and in the global attention process in the vision transformer layers. It should be noted here, compared to the image generator that is always the same between our method and existing methods, the motion estimator tends to take a small computation cost in the whole image animation process. While being considerably lighter than the image generator, the motion estimator is proved to be efficient and effective for improving image animation performance, as in our method and recent works~\cite{siarohin2020first,PCAMotion}. 

\begin{figure*}[ht]
\centering
  \includegraphics[width=1\linewidth]{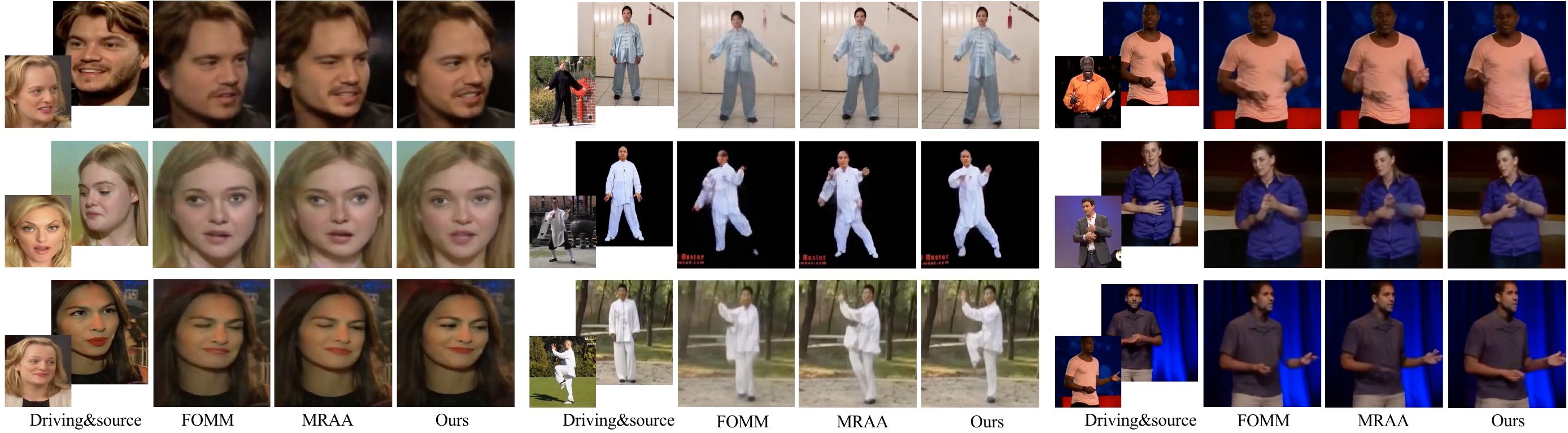}
  \caption{Qualitative comparisons on the cross-identity image animation task. We show results on three datasets (from left to right: VoxCeleb, TaichiHD and TEDTalks), each with three paired examples.}
  \label{fig:animation compare}
\end{figure*}

\noindent\textbf{Quantitative comparison:} The video reconstruction results are presented in Table~\ref{tab1}. As can be seen from the table, our method generally performs the best across all evaluation metrics, as well as across all benchmark datasets with object types including human body, human face and animal etc., reflecting the superiority of the motion transformer to perform general image animation. More specifically, a lower L1 distance straightforwardly indicates better video reconstruction quality achieved by our method. 
\begin{table}[]
\caption{Parameters comparison of the proposed motion estimator with that of FOMM and MRAA. For clearness, we also list the parameters of the image generator (the encoder-decoder generator together with the mask predictor). The model parameters and FLOPs are computed with input image resolution $256\times256$. }
\label{params_gflops}
\centering
{
\begin{tabular}{ccc}
\toprule[1.0pt]
\multicolumn{1}{c|}{}                        & Parameters    & FLOPs \\ \hline
\multicolumn{1}{l|}{ImageGenerator}               & 45.57M         & 53.64G  \\
\multicolumn{1}{l|}{MotionEstimator-FOMM}    & 14.21M        & 1.28G   \\
\multicolumn{1}{l|}{MotionEstimator-MRAA}    & 14.20M        & 1.26G   \\
\multicolumn{1}{l|}{MotionEstimator-Ours}    & 12.23M        & 7.54G  \\
\bottomrule[1.0pt]
\end{tabular}
}
\end{table}
\begin{table*}[]
\caption{Quantitative comparisons with FOMM~\cite{siarohin2020first} and MRAA~\cite{PCAMotion} on the video reconstruction task. We present results on four benchmarks, our method generally achieves the best performance on all datasets across all metrics.}
\label{tab1}
\centering
\resizebox{1.0\textwidth}{!}
{
\begin{tabular}{c|ccc|ccc|ccc|c}
\toprule[1.0pt]
\multicolumn{1}{l|}{} & \multicolumn{3}{c|}{TaiChiHD}                                    & \multicolumn{3}{c|}{TEDTalks}                                     & \multicolumn{3}{c|}{VoxCeleb}             & MGIF \\
\multicolumn{1}{l|}{} & L1             &  (AKD, MKR)               & AED                  & L1     &  (AKD, MKR)      & AED                                     & L1        & AKD         & AED              & L1   \\ \hline
FOMM                  & 0.057          &  (6.649, 0.036)           & 0.172                & 0.029  &  (4.382, 0.008)   & 0.127                                  & 0.041     & 1.29       & 0.133            & 0.0224\\
MRAA              & 0.048              &  (5.246, 0.024)           & 0.150                & 0.027  &  (3.955, \textbf{0.007}) & 0.118                           & 0.040     & 1.28       & 0.133            & 0.0274\\
Ours             & \textbf{0.045} &  (\textbf{4.670, 0.021}) &\textbf{0.148}      & \textbf{0.026} &  (\textbf{3.456, 0.007}) & \textbf{0.113}          & \textbf{0.038} & \textbf{1.18} & \textbf{0.116} & \textbf{0.0200}\\
\bottomrule[1.0pt]
\end{tabular}
}
\end{table*}
It is also worth noting that our method achieves considerable improvements in terms of AKD on the three datasests, which strongly suggests that our method achieves better transferred motion. This can be further validated in the qualitative results of Fig.~\ref{fig:animation compare}. Moreover, our method also achieves the best performance on the AED metric, indicating that the identity information can be better preserved using our method for conducting image animation. The superiority on the AED metric is even more obvious on the VoxCeleb dataset, we draw reason that the identity information is especially important for a human face, while our method generally learns the global motion pattern for the human face, which enables it to better capture the global face structure.

\noindent\textbf{User preference:} To evaluate the cross-identity image animation, we conduct a user study with fifty participants. In more detail, we first prepare fifty comparison videos, each of which is a concatenation of a source image, a driving video featuring a different-identity, and videos generated by the three methods. Note that the spatial locations of the generated videos are randomly placed. Participants are required to evaluate these three videos according to the transferred motion and identity preservation. The results in Table~\ref{tab2} show that our method is clearly awarded more user preferences than other existing methods. 

\noindent\textbf{Qualitative comparison:} In Fig.~\ref{fig:animation compare}, we present representative animation examples on the TaichiHD, Voxceleb1 and TEDTalks dataset. As the figure shows, our method is generally better at handling both global and local motions. In more detail, for the human face, despite the fact that both FOMM and MRAA can capture the head rotation, our method can synthesize the most realistic and detailed expression information. 

Our analysis suggests that it is often the case that a rigid human face turns from left to right, and occlusion occurs in this kind of rigid or global motion. Our method can effectively learn global motion patterns for a human face; 
\begin{table}[]
\begin{minipage}[t]{0.482\linewidth}
\caption{User preferences of our method against FOMM and MRAA on the TaichiHD, TEDTalks, and Voxceleb dataset.}
\label{tab2}
\centering
{
\begin{tabular}{cccc}
\toprule[1.0pt]
\multicolumn{1}{l|}{}            & TaiChiHD    & TEDTalks     & VoxCeleb   \\ \hline
\multicolumn{1}{c|}{FOMM}        & 96.5\% & 66.4\% & 60.8\% \\
\multicolumn{1}{c|}{MRAA}    & 68.5\% & 57.1\% & 69.8\% \\
\bottomrule[1.0pt]
\end{tabular}
}
\end{minipage}
\begin{minipage}[t]{0.482\linewidth}
\caption{Performance comparison on the TaichiHD dataset with and without position encoding, denoted as $w$ PE and $w/o$ PE.}
\label{PE ablation}
\centering
\begin{tabular}{c|ccc}
\toprule[1.0pt]
\multicolumn{1}{l|}{}             & L1               &  (AKD, MKR)               & AED                     \\ \hline
$w/o$ PE                                 & 0.047            &  (5.482, 0.028)           & 0.158                   \\
$w$ PE                                & \textbf{0.045}   &  (\textbf{4.670, 0.021})  & \textbf{0.148}                   \\
\bottomrule[1.0pt]
\end{tabular}
\end{minipage}
\end{table}
accordingly, this makes it easier to detect the occlusion caused by the head rotation and then guide the image generator to inpaint this occluded face structure. 
In FOMM and MRAA, the motion is learned in a relatively local manner, which makes it more difficult to capture the global face structure. For the human body, it can also be observed that our method synthesizes the most motion-stable results, while FOMM and MRAA often fail to capture the driving motions. We believe this occurs because, lacking awareness of the global motion information, FOMM and MRAA are easier to be affected by large motions and intervention of background features. By contrast, our motion transformer learn the part motions in a global-assisted fashion, enabling it to learn the more stable part motions.

\subsection{Ablation Study and Parameter Analysis}

In this section we study the influence of different components of our motion transformer. More specifically, we conduct video reconstruction experiments on the TaichiHD dataset for the purpose of quantitative analysis.

\noindent\textbf{Position encoding:} As can be seen Table~\ref{PE ablation}, it is crucial to add position encoding to the image tokens in our experiments. We can explain the importance of position encoding from two angles. On one hand, the motion (\aka, keypoint and its corresponding affine matrix) estimation is a highly position-sensitive task, in which image tokens equipped with position encoding ease the learning process. On the other hand, in order to learn geometry-consistent keypoint and affine matrix representations in an unsupervised manner, 
\begin{table}[]
\begin{minipage}[t]{0.482\linewidth}
\centering
\caption{Performance comparison on the TaichiHD dataset with different CNN backbones of the motion transformer.} 
\label{CNN ablation}
\resizebox{1.0\linewidth}{!}{
\begin{tabular}{c|cccc}
\toprule[1.0pt]
\multicolumn{1}{c|}{CNN}    &Param.           & L1               &  (AKD, MKR)               & AED                     \\ \hline 
Stem                        &5.56M             & 0.048            &  (6.056, 0.030)           & 0.161                   \\
HR-W32                      &12.23M           & \textbf{0.045}   &  (\textbf{4.670}, 0.021)  & \textbf{0.148}                  \\
HR-W48                      &21.30M           & \textbf{0.045}   &  (4.829, \textbf{0.020})  & 0.149                  \\
\bottomrule[1.0pt]
\end{tabular}
}
\end{minipage}
\hphantom{F}
\begin{minipage}[t]{0.482\linewidth}
\centering
\caption{Performance comparison on the TaichiHD dataset with respect to different numbers of transformer layers.}
\label{tab3}
\resizebox{0.80\linewidth}{!}{
\begin{tabular}{c|ccc}
\toprule[1.0pt]
\multicolumn{1}{c|}{Layers}             & L1               &  (AKD, MKR)               & AED                     \\ \hline
4                                 & 0.046            &  (5.320, 0.027)           & 0.155                   \\
8                                 & 0.046            &  (5.226, 0.025)           & 0.154  \\
12                                & \textbf{0.045}   &  (\textbf{4.670, 0.021})  & \textbf{0.148}                   \\
\bottomrule[1.0pt]
\end{tabular}
}
\end{minipage}
\end{table}
the equivariance loss (\ie, Eqn.~(\ref{equi loss})) is considered to be more effective with position encoding, since the order of the image patches are shuffled by a geometric transformation; however, the position encoding is invariant to the shuffling process, thus the consistency loss can enforce the network to better capture useful image patch features.

\noindent\textbf{CNN encoder:} To explore the influence of the image feature representation, we implement the motion transformer with different CNN backbones. As can be seen from Table~\ref{CNN ablation}, compared to our basic setting (\ie, HR-w32), a light-weight CNN (\ie, Stem net~\cite{sun2019deep}, a widely used CNN for quickly down-sampling the image by a scale factor of 4) yields worse performance, while a heavy CNN (\ie, HR-w48~\cite{sun2019deep}) brings no significant improvement. We accordingly conclude that in the absence of a good image feature representation, the motion transformer can't work well; at the same time, the promotion of the CNN backbone to the final performance is limited, which we attribute to the lack of supervision in the unsupervised image animation.

\noindent\textbf{Vision transformer layers:} We further conduct experiments to explore the influence of different numbers of vision transformer layers used in our motion transformer. As can be seen from Table~\ref{tab3}, with smaller numbers of layers, the performance declines considerably (especially on AKD and MKR, which evaluate the motion quality). This reflects the fact that the motion transformer with relatively deeper transformer layers facilitates to learn better motion embeddings for the regression of the motion information.

\subsection{Visualization}

In this section, we visualize intermediate results to analyze how the motion transformer learns global motions with different object types and what motion patterns have been learned. Samples are randomly chosen; while our observations suggest that the model behaves similarly on the entire dataset.

\begin{figure*}[ht]
\centering
  \includegraphics[width=1\linewidth]{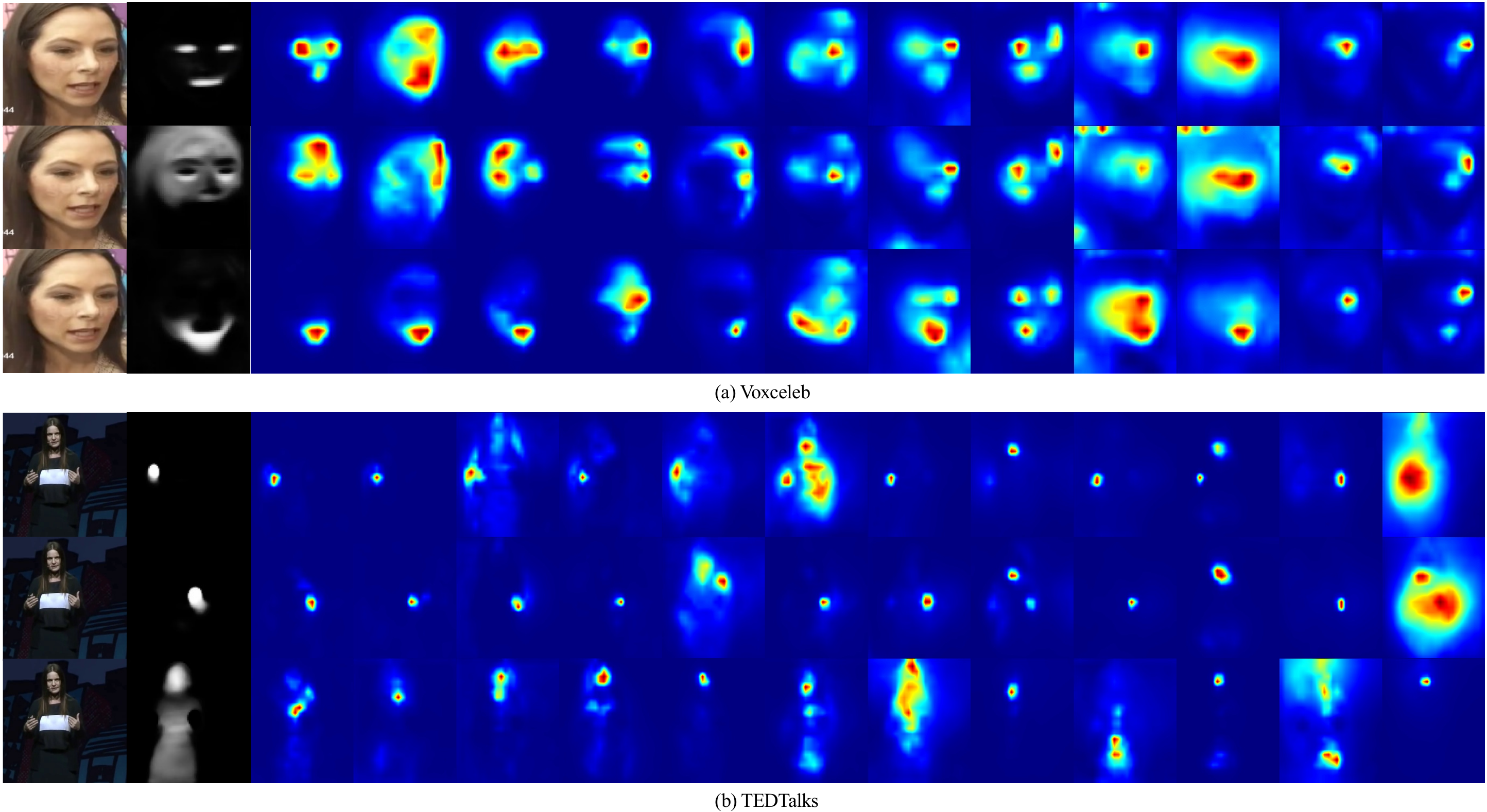}
  \caption{Visualizations of visual attention maps between motion tokens and image tokens on the VoxCeleb  and TEDTalks datasets. From left to right of in each row, the presented content respectively represents the driving image, the corresponding motion mask and the visual attention maps of each transformer layer. We present visual attention maps of three representative motion tokens for each dataset. Note that we reshape and resize the sequence attention values to the original image sizes.}
  \label{fig:cross attention}
\end{figure*}

\noindent\textbf{Visual attention:} We visualize the attention maps between motion tokens and image tokens, to reveal how the motion transformer learns the global and local motion. The results on the VoxCeleb and TEDTalks dataset are presented in Fig.~\ref{fig:cross attention}. For human faces, it can be seen that the whole face region tends to be attended by all different motion tokens;
this implies that each motion part is learned with awareness of the global motion, which is in line with our motivation. For human bodies, we first observe that the global motion is effectively captured; as can be seen in the third row, the motion token learned with a global pattern attends almost the whole regions of the object in the image. Moreover, we find that in the initial transformer layers, the local motions of the woman's hands are well captured. As the depth increases, motion tokens can also find some other meaningful relationship. For example, as illustrated in the attention results of the TEDTalks sample in Fig.~\ref{fig:cross attention}, the motion token representing the woman's right hand (\ie, the first row) actually shows that it is related to her upper body and head, as in the attention map of the sixth transformer layer. This is reasonable because, when a woman presents something in the talk, her body may move together with the hand gestures to communicate with the whole body language. 





\section{Conclusion}
We propose a new method, called the motion transformer, under the general image animation framework for unsupervised image animation. The motion transformer introduces both image tokens and learnable motion tokens. To encourage the interactions between image and motion tokens, our motion transformer network employs multiple transformer layers, which take those tokens as input in order to learn the underlying motion relationship and obtain better motion embeddings. We further conduct extensive experiments on four benchmark datasets. Our experimental results validate the effectiveness of capturing the global motion information in our motion transformer.

\noindent\textbf{Acknowledgement:} This work is supported by the Major Project for New Generation of AI under Grant No. 2018AAA0100400, the National Natural Science Foundation of China (Grant No. 62176047), Sichuan Science and Technology Program (No. 2021YFS0374, 2022YFS0600), Beijing Natural Science Foundation (Z190023), and Alibaba Group through Alibaba Innovation Research Program. This work is also partially supported by the Science and Technology on Electronic Information Control Laboratory.


\clearpage
%
%
\bibliographystyle{splncs04}
\bibliography{egbib}
\end{document}